\documentclass[runningheads]{llncs}
\usepackage[T1]{fontenc}
\usepackage{graphicx}
\usepackage{booktabs}
\usepackage[misc]{ifsym}
\newcommand{\corr}{(\Letter)}
\usepackage{mwe}
\usepackage{adjustbox}
\usepackage{amsmath, amssymb}
\begin{document}

\title{Using Subgraph GNNs for Node Classification: \\an Overlooked Potential Approach}

\titlerunning{Using Subgraph GNNs for Node Classification}
% If the full title of your paper is short enough to also fit in the running head, you can omit the abbreviated paper title here. You can check as follows: if you comment out the \titlerunning line, something will appear in the header of all odd-numbered pages of your PDF from page 3 onward. This something is either the full title (in which case all is well), or the error message "Title Suppressed Due to Excessive Length". If this error message appears, you're going to want to provide an abbreviated title within the \titlerunning command, because if you won't do it, Springer will do it for you.

%N.B.: Author information (both in the \author{} and \authorrunning{} command) should only be present in the Camera-Ready Version of your paper. The version that you initially submit for review, ought to be double-blind. So, when initially submitting your paper, use:
% \author{Author information scrubbed for double-blind reviewing}
\author{Qian Zeng\inst{1}\and
Xin Lin\inst{1}\corr \and
Jingyi Gao \inst{1}\and
Yang Yu\inst{1}}
\authorrunning{Q. Zeng et al.}
% \author{Andr\'e Lauren Benjamin\inst{1} \and
% Calvin Cordozar Broadus Jr.\inst{2,3} \corr \and
% Antwan Andr\'e Patton\inst{1}\orcidID{0000-1111-2222-3333}}
% You may leave out the orcidID information, if you want to.
% Use \corr to indicate the corresponding author. Note the spacing around the \corr command. Only one author can be the corresponding author.

%N.B.: comment out the \authorrunning{} command for the double-blind version of your paper submitted for review. Later, if your paper is accepted, use the command for the Camera-Ready Version.
% \authorrunning{A.L. Benjamin et al.}
% If there are two authors, write 'A.L. Benjamin and C.C. Broadus Jr.'
% If there are more than two authors, '[...] et al.' is used.
\institute{East China Normal University, Shanghai, China
\email{xlin@cs.ecnu.edu.cn,\{qzeng,jygao,yuyang\}@stu.ecnu.edu.cn}}
% \institute{Fictional Southern University, Savannah GA 31404, USA \email{\{a.l.benjamin,a.a.patton\}@fsu.fake}
% \and
% Fictional West Coast University, Long Beach CA 90840, USA \email{ccb@fwcu.fake}
% \and
% Secondary European Affiliation, Tiergartenstr. 17, 69121 Heidelberg, Germany
% \email{lncs@springer.com}}

\maketitle              % typeset the header of the contribution

\begin{abstract}
Previous studies have demonstrated the strong performance of Graph Neural Networks (GNNs) in node classification. However, most existing GNNs adopt a node-centric perspective and rely on global message passing, leading to high computational and memory costs that hinder scalability. To mitigate these challenges, subgraph-based methods have been introduced, leveraging local subgraphs as approximations of full computational trees. While this approach improves efficiency, it often suffers from performance degradation due to the loss of global contextual information, limiting its effectiveness compared to global GNNs. To address this trade-off between scalability and classification accuracy, we reformulate the node classification task as a subgraph classification problem and propose SubGND (Subgraph GNN for NoDe). This framework introduces a differentiated zero-padding strategy and an Ego-Alter subgraph representation method to resolve label conflicts while incorporating an Adaptive Feature Scaling Mechanism to dynamically adjust feature contributions based on dataset-specific dependencies. Experimental results on six benchmark datasets demonstrate that SubGND achieves performance comparable to or surpassing global message-passing GNNs, particularly in heterophilic settings, highlighting its effectiveness and scalability as a promising solution for node classification.

\end{abstract}

\section{Introduction}
% \subsection{Main Contributions}
Previous research has demonstrated the effectiveness of Graph Neural Networks (GNNs) across a wide range of graph learning tasks, including node classification, link prediction, and graph classification. By leveraging efficient message-passing mechanisms\cite{gilmer2017neural}, GNNs integrate information from neighboring nodes to learn meaningful representations. However, most existing GNN architectures rely on global message passing, which requires multiple rounds of information exchange across the entire graph to achieve optimal performance. This dependency introduces several limitations: 

\textit{First}, it assumes that both the training and test sets exist within the same static graph, restricting the model’s applicability in dynamic or incremental graph scenarios. \textit{Second}, it incurs high computational and memory overhead, making it impractical to scale to large graphs with billions of nodes and edges, especially in resource-constrained environments. \textit{Finally}, similar to global message-passing GNNs, it also encounters over-smoothing \cite{oonograph} issues, which degrade the model's ability to preserve discriminative node representations. 

Such limitations hinder the broader adoption and application of GNNs. To address these challenges, subgraph-based inductive methods, such as GraphSAGE\cite{hamilton2017inductive} and GraphSAINT\cite{zeng2019graphsaint}, have been proposed. These methods sample local subgraphs for each node, perform message passing within them, and employ models like GCN\cite{kipf2016semi} or GAT\cite{velivckovic2017graph} for representation learning. By restricting computation to local neighborhoods, they enhance scalability and adaptability while significantly reducing complexity compared to full-graph GNNs. Consequently, subgraph-based approaches are increasingly regarded as a more efficient and flexible solution for large-scale graph learning.

However, existing subgraph-based methods still face critical challenges. Their reliance on limited neighborhood sampling leads to inherent information loss, creating a performance gap compared to global message-passing approaches. While techniques such as Local Message Compensation (LMC\cite{shilmc}) attempt to mitigate this issue by recovering missing information and correcting gradient bias, they introduce additional computational overhead and require specialized module adjustments. Moreover, although these methods perform well on homophilic graphs, they struggle to generalize effectively in heterophilic settings due to structural and feature disparities among neighboring nodes. 

Nevertheless, the scalability advantages of subgraph-based methods position them as a promising avenue for further research. Therefore, this work aims to develop a subgraph-based node classification framework that matches the performance of global message-passing GNNs while exhibiting superior generalization in heterophilic graphs. To this end, we reformulate the node classification problem as a graph classification task and propose SubGND (Subgraph GNN for NoDe), a novel framework designed to adapt this formulation. Specifically, we first identify the issue of label conflicts arising during subgraph transformations and introduce a Differentiated Zero-Padding strategy alongside an Ego-Alter subgraph representation to mitigate this problem and enhance model expressiveness. Additionally, given that the reliance on structural information varies across datasets, we introduce an Adaptive Feature Scaling Mechanism that dynamically adjusts feature contributions based on dataset-specific dependencies. Experimental results demonstrate that SubGND achieves performance comparable to global GNNs in homophilic graphs while significantly outperforming existing methods in heterophilic settings, validating its effectiveness.

In summary, the main contributions of this paper are as follows:
\begin{itemize}
    \item To the best of our knowledge, this work is the first to explore node classification entirely from a subgraph perspective, introducing a novel approach to applying GNNs to node classification through problem reformulation.
    \item We identify the label conflict issue commonly encountered during subgraph transformation and propose a novel subgraph framework that includes a Differentiated Zero-Padding strategy, a distinct Ego-Alter subgraph representation method, and an Adaptive Feature Scaling Mechanism.
    \item Extensive experiments on multiple public datasets demonstrate that SubGND achieves performance comparable to or exceeding existing node-based GNN models, validating its effectiveness across different scenarios.
\end{itemize}

\section{Related Work}
\subsubsection{Subgraph-Based GNNs.} Research on subgraph-based methods for node classification can generally be categorized into two main approaches:

The first category extends global message-passing models by leveraging subgraph sampling to enhance scalability. These methods construct sampled subgraphs as computational trees for nodes while employing models such as GCNs \cite{kipf2016semi} and GATs \cite{velivckovic2017graph}. The focus is on improving sampling strategies to balance efficiency and information retention. For example, GraphSAGE \cite{hamilton2017inductive} approximates global information by sampling fixed-size neighbor subsets, while GraphSAINT \cite{zeng2019graphsaint} designs multiple sampling strategies to ensure unbiasedness. Additionally, LMC \cite{shilmc} introduces local message compensation to reduce performance gaps with global message-passing GNNs. However, these methods remain node-centric, primarily aiming to bridge the gap with global message-passing models and still struggle in heterophilic graphs.

The second category explores pretraining-based approaches, unifying node and graph classification tasks. GraphPrompt \cite{liu2023graphprompt} and Self-Pro \cite{gong2023prompt} reframe node classification as prototype subgraph similarity matching, while ProG \cite{sun2023all} applies meta-learning for prompt-based learning. GCC \cite{qiu2020gcc} employs contrastive learning to enhance node representation. These works establish theoretical connections between node and subgraph classification but remain primarily within the pretraining domain, lacking direct application to node classification tasks.

While subgraph methods are efficient for large-scale datasets and related to node classification, no existing work fully adopts a subgraph-centric reformulation, leaving an open research direction.

\subsubsection{Heterophilic GNNs.} Recent advancements in addressing graph heterogeneity can be categorized into several key approaches. The most common approach focuses on capturing varied neighborhood relationships, as exemplified by MixHop \cite{abu2019mixhop}, which mixes feature representations from neighbors at various distances to address graph heterogeneity. GCNII \cite{chen2020simple} tackles heterogeneity in deeper networks through initial residuals and identity mappings, aiming to capture diverse patterns. Another approach emphasizes improving feature aggregation by decoupling processes, such as in FSGNN \cite{maurya2022simplifying}, which decouples feature aggregation from network depth and uses softmax for feature selection, and in H2GNN\cite{zhu2020beyond}, which separates ego and neighbor embeddings while leveraging higher-order neighborhoods. Additionally, adaptive mechanisms are employed by methods like ACM-GCN \cite{luan2022revisiting}, which uses adaptive channel mixing to extract richer localized information, and GPR-GNN \cite{chien2020adaptive}, which learns Generalized PageRank weights to optimize the integration of node features and topology. Moreover, GRAD.GATING \cite{rusch2023gradient} and GloGNN++ \cite{li2022finding} focus on global and multi-rate information processing, with GRAD.GATING introducing gradient modulation to counteract over-smoothing, and GloGNN++ aggregating global information through a learned coefficient matrix to effectively capture node correlations.

Notably, all these models are node-centric approaches. Since there are no comparable models that address the problem from a subgraph classification perspective as we do, we primarily conduct comparisons in this paper against the aforementioned node-centric methods.

\section{Preliminaries}
\textbf{Problem Definition.} Consider a graph \( G = (V, E) \) where \( V \) denotes the set of nodes and \( E \subseteq V \times V \) represents the edges. Each node \( v \in V \) is initially characterized by a feature vector \( \mathbf{x}_v \in \mathbb{R}^d \). 

For each node \(v\), we obtain its multi-hop neighborhood \(N_v\) and the set of edges between these nodes through appropriate sampling techniques. The subgraph composed of this set of nodes and edges is referred to as the induced subgraph of node \(v\). In this context, node \(v\) is termed the \textbf{ego-vertex} of the subgraph, while the other nodes are referred to as the \textbf{alter-vertex}. Our objective in this paper is twofold: 

First, to devise appropriate strategies to construct subgraphs \( G_v = (V_v, E_v) \) for each node \( v \); 

Second, to learn a function \( g : \mathbb{R}^{|V_v| \times d} \to \mathcal{Y} \) that maps each subgraph \( G_v \) to a label \( y_v \), aligning with the label of the central node \( v \).For each node \( v \), \( G_v \) is constructed by including \( v \) and its neighboring nodes (i.e alter-vertex) within a predefined radius \( r \). The edges \( E_v \subseteq E \) are defined accordingly. The task then involves minimizing the classification error, formulated as:
\begin{equation}
\min_g \sum_{v \in V} \mathcal{L}(g(G_v), y_v)
\end{equation}
where \( \mathcal{L} \) quantifies the disparity between the predicted label \( g(G_v) \) and the true label \( y_v \).

Thus, the challenge encompasses both the strategic construction of informative subgraphs \( G_v \) and the development of an effective mapping function \( g \) to achieve accurate subgraph classifications.

\section{Methodology}

\begin{figure*}[t]
\centering
\includegraphics[width=\textwidth]{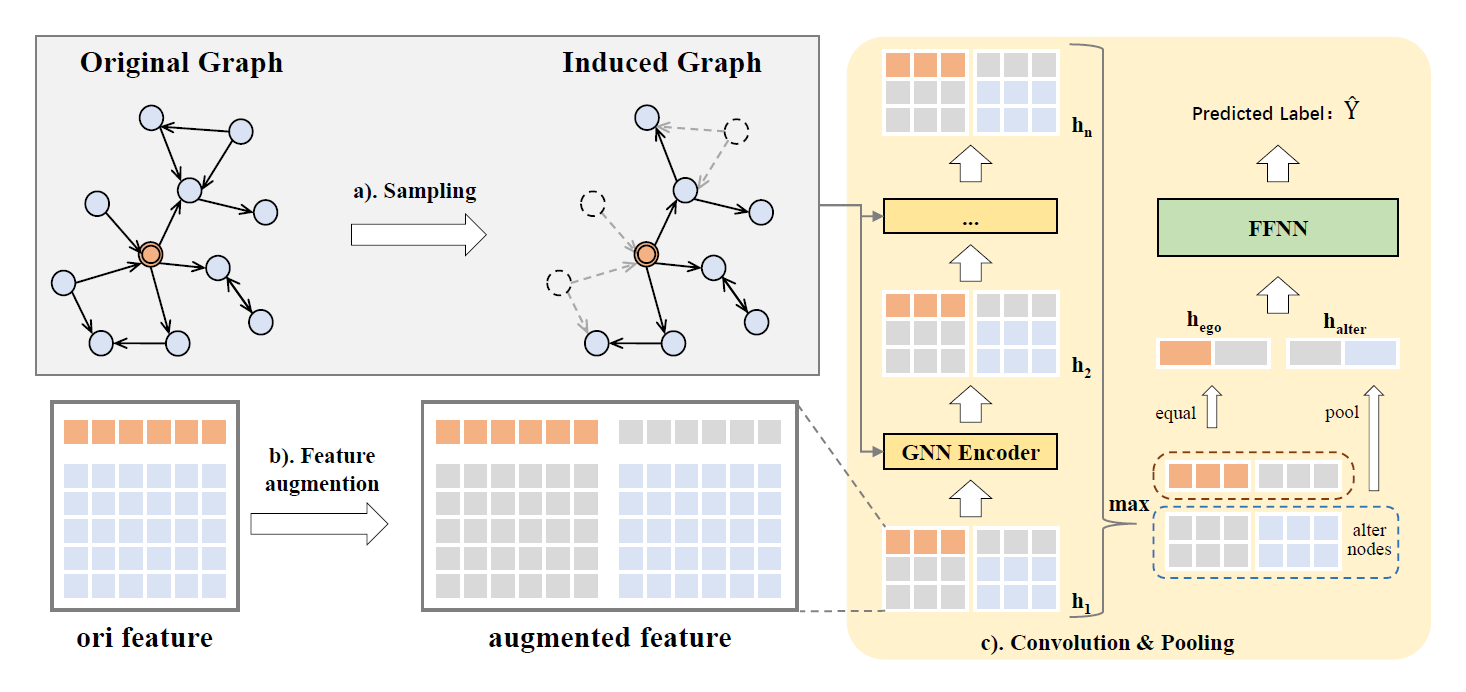} 
\caption{Overall Framework: a) Induced subgraphs are constructed using a slightly modified variant of the ISRW algorithm, with each subgraph labeled by the ego-vertex's label. b) Node representations are augmented by distinct zero-padding operations are applied separately to the ego-vertex and alter-vertex. c) Finally, node embeddings are updated via a GNN encoder, and the ego-vertex embedding is concatenated with pooled representations of alter-vertex set for the final prediction.}
\label{fig2}
\end{figure*}

Converting a node classification task into a subgraph classification task typically involves two key steps: \textit{Induced Graph Generation}, followed by \textit{Subgraph Representation and Prediction}. We will introduce our approach in alignment with these steps, and the brief summaries are as follows:

In the first step, we utilize the ISRW algorithm, which provides greater flexibility and control than traditional neighbor sampling methods \cite{hamilton2017inductive}, to generate induced subgraphs for nodes. Additionally, we introduce adaptive modifications based on task requirements, such as adjusting the timing of bidirectional edge conversion and implementing a dual-sampling strategy to enhance sampling stability.

In the second step, we introduce a differentiated zero-padding strategy that adds zero vectors at specific positions in node representations to enhance expressiveness. We further employ an Ego-Alter subgraph representation method to address label conflict issues, particularly in heterophilic graphs. Finally, we apply an Adaptive Feature Scaling Mechanism to adjust feature importance across different datasets. 

Overall, our framework reformulates the task through the aforementioned steps and efficiently integrates subgraph information with individual node characteristics, ultimately achieving performance comparable to or surpassing state-of-the-art methods across multiple datasets, thereby demonstrating its potential.

\subsection{Induced Graph Generation}
\subsubsection{Graph Sampling.} Graph sampling is a technique used to extract representative subgraph samples from the original graph. In this paper, we adopt the graph sampling approach mentioned in GCC \cite{qiu2020gcc} to create induced subgraphs. Specifically, the process involves three primary steps.\footnote{For further details, please refer to the original text in \cite{qiu2020gcc}. Additionaly, Unlike GCC, which uses position embeddings as the primary features, we include the original node features} :

\begin{itemize}
    
    \item \textit{Random Walk with Restart}: Initiate an iterative random walk on graph \( G \) from vertex \( v \), with each step having a probability of returning directly to \( v \) at each step.
                    
    \item \textit{Subgraph Induction}: Retrieve the edges corresponding to the set of nodes obtained from the first step. \footnote{It is also known as Induced Subgraph Random Walk Sampling (ISRW).}
    
    \item \textit{Anonymization}: Relabel each subgraph's vertices to \( \{1, 2, \ldots, |V_{S_v}|\} \) in arbitrary order. In addition, we retain the corresponding node features and labels for subsequent processing.
    
\end{itemize}

\subsubsection{Directed vs Undirected.} Previous works, such as Digraph Transformer \cite{geisler2023transformers} and Dir-GNN \cite{rossi2024edge}, have demonstrated the significant value of edge directionality in directed graphs. In the context of sampling processes for directed graphs (note that in undirected graphs, there is no distinction among these aspects), two key steps are influenced by edge direction. First, during the random walk, the choice of neighbors—whether they are incoming, outgoing, or both—directly affects the size and scope of the induced subgraphs. Second, the decision to add bidirectional edges before message passing is crucial for effective information propagation. 
In our experiments, we found that converting to bidirectional edges is necessary for effective message propagation, but this should be done after induced-subgraph construction. Walking with out-neighbors during the induced-subgraph construction phase yielded better results. Therefore, we default to using out-neighbors for the random walk during induced-subgraph construction and convert to bidirectional edges afterward.

\subsubsection{Dual-Sampling Strategy.} To mitigate the adverse effects of randomness from single-sample procedures on model predictions, we adopt a dual-sampling approach for each node, followed by independent gradient computations for each sample. Empirical results demonstrate that this method effectively reduces sampling variability, stabilizes the training process, and enhances model generalization. \footnote{Additional sampling iterations were tested, but they yielded no substantial performance gains relative to the increased resource cost, affirming the efficiency of the two-sample approach.}

\subsubsection{Crucial Hyperparameters.} $restart\_probability$ and $rw\_hops$ \footnote{essentially $max\_nodes\_per\_seed$, renamed for consistency with neighbor sampling} are two important parameters in the sampling process. The former governs the trade-off between local and global information by controlling how deep or broad the random walk extends. A higher restart probability leads to smaller, more focused subgraphs that emphasize local details, while a lower value allows for a broader exploration of the network, which is crucial for tasks requiring a global view. $rw\_hops$ limits the number of nodes visited, affecting the size and diversity of the subgraphs generated. In our work, we follow the GCC \cite{qiu2020gcc} framework by setting the restart probability to 0.8 and treat $rw\_hops$ as a flexible hyperparameter to fine-tune the subgraph generation process.

\subsubsection{Dead-End Nodes} To ensure consistency with other nodes, we define dead-end nodes as those without available neighbors for the random walk—meaning they lack either outgoing or incoming connections depending on the walk direction. These nodes are augmented with self-loops and treated as distinct subgraphs.

\subsection{Subgraph Representation and Prediction}

\subsubsection{Backbone.} Unlike prior models that rely on GCN or GAT for node representation learning, we adopt the Graph Isomorphism Network (GIN) \cite{xu2018powerful} as the backbone, which has been shown to be more expressive in capturing structural distinctions. The core operations of this model are as follows:
\begin{equation}
\label{eq2}
h_v^{(k)} = \text{MLP}^{(k)} \left( (1 + \epsilon^{(k)}) \cdot h_v^{(k-1)} + \sum_{u \in N(v)} h_u^{(k-1)} \right)
\end{equation}

Here, \( h_v^{(k-1)} \) represents the node \( v \)'s representation at the previous \( k-1 \) layer, \( \epsilon^{(k)} \) is a learnable parameter \footnote{In this paper, we treat it as a hyperparameter for optimization, as discussed in the subsequent sections}, \( N(v) \) denotes the neighbors of node \( v \), and \( \text{MLP}^{(k)} \) refers to a multi-layer perceptron for (non-)linear mapping.And then use a readout function to aggregates node representations \( h_v^{(K)} \) across all nodes \( v \) in the graph to derive a comprehensive graph-level representation \( h_G \)

Through iterative updates at each layer, the GIN model effectively captures node information and intricate graph structures, demonstrating strong performance in tasks such as node and graph classification.
\begin{figure*}
\centering
\includegraphics[width=\textwidth]{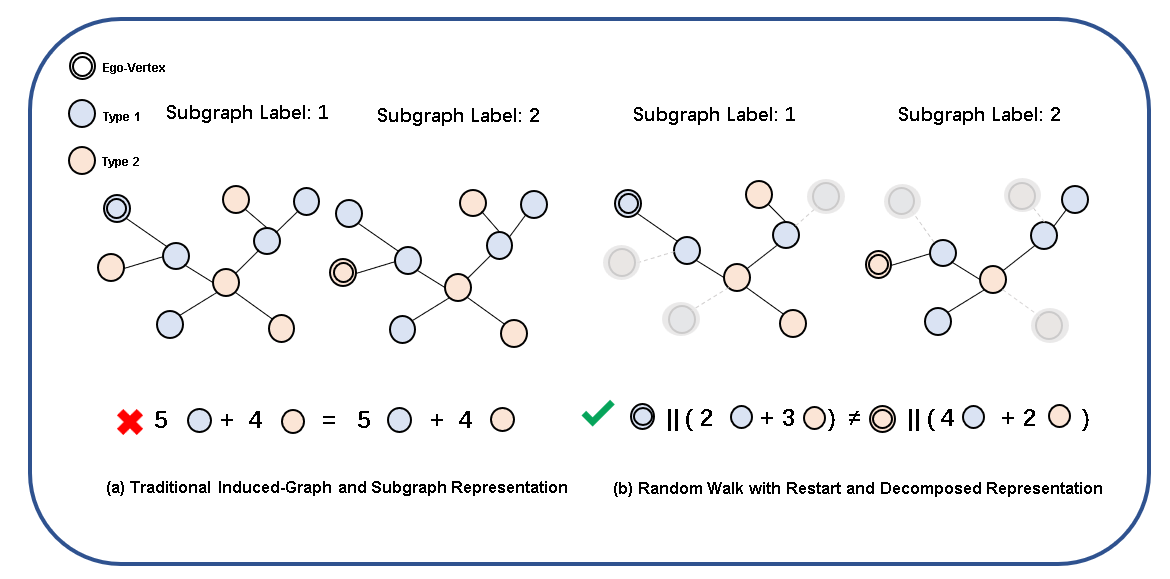} % Reduce the figure size so that it is slightly narrower than the column.
\caption{(a) Different types of ego-vertex  may possess identical neighboring subgraphs, leading to the introduction of conflicting information into the model. (b) Through the application of Random Walk with Restart (RWR) and decomposed representations, We can gain more differentiated perspectives to alleviate this problem }
\label{fig1}
\end{figure*}
\subsubsection{Differentiated Zero-Padding.} Applying a GIN model directly to induced subgraphs can lead to significant challenges, namely label conflict problems. As illustrated in figure 1, two starting nodes with identical neighboring subgraphs require distinct subgraph labels— 1 and 2 — owing to the differing types of the starting nodes. The original model encounters difficulty in distinguishing between such subgraphs. To mitigate this issue, we first employ the ISRW algorithm, as outlined in earlier sections, to minimize the occurrence of these scenarios. Additionally, we decouple the representation of the ego-vertex from that of the surrounding nodes. 

Specifically, we implement a differentiated zero-padding strategy for the node representations within each subgraph. For the ego-vertex, we append a zero vector of equivalent dimensionality to the end of its representation, while for the remaining nodes, the zero vector is prepended to the beginning. The expression is given as:

\begin{equation}
\mathbf{z}_i = 
\begin{cases} 
\text{Concat}(\mathbf{x}_i, \mathbf{0}) & \text{if } i = v \\
\text{Concat}(\mathbf{0}, \mathbf{x}_i) & \text{if } i \neq v
\end{cases}
\end{equation}

where \(\mathbf{z}_i\) denotes the augmented feature vector for node \(i\) within the subgraph \(g\), and \(v\) is the ego-vertex. \footnote{To prevent an excessive increase in dimensionality, this operation is performed only after the initial linear mapping to the hidden size.}

This differentiated operation effectively embeds unique identity information into each node, ensuring that even nodes within identical structures exhibit variance in their representations depending on the starting node. Furthermore, this strategy enhances the model’s ability to capture structural nuances, a concept validated in previous works such as ID-GNN \cite{you2021identity}, which labels each node with a one-hot identity encoding, and GLASS \cite{wang2021glass}, which employs a max-zero-one labeling technique. The key distinction in our approach lies in its heightened emphasis on the starting node's influence on the surrounding nodes.

\subsubsection{Ego-Alter Subgraph Representation.}
After applying a specialized feature augmentation strategy, we first perform convolution on each node using a basic GIN model to obtain the node representations at each layer as follows:

\begin{equation}
h_{v_i}^{(l)} = \text{GIN}^{(l)}\left(h_{v_i}^{(l-1)}, \text{Agg}_{v_j \in \mathcal{N}(v_i)}(h_{v_j}^{(l-1)})\right)
\end{equation}

where \( h_{v_i}^{(l)} \) denotes the representation of node \( v_i \) at layer \( l \), and \(\text{Agg}\) is the aggregation function used by the GIN model.

For each node, considering the diversity of information across different layers, we employ MaxPooling to aggregate representations from multiple layers, thereby enhancing node features. For individual subgraphs, we perform a pooling operation (such Max, Mean or Sum) on alter node features and concatenate the pooled result with the central node’s features to obtain the final subgraph representation. The formulation is as follows
\begin{equation}
\label{eq:34}
h_v^{\text{final}} = \text{MaxPooling}(h_v^{(1)}, \dots, h_v^{(L)})
\end{equation}
\begin{equation}
\label{eq:35}
h_{\text{sub}} = \left[ h_{v_{\text{ego}}}^{\text{final}} \parallel \text{Pooling}\left( \{ h_{v_{\text{alter}}}^{\text{final}} \mid v_{\text{alter}} \in V_{\text{alter}} \} \right) \right]
\end{equation}

Through this approach, the subgraph representation integrates the enhanced features of the central node with the aggregated features of its neighboring nodes. This representation not only emphasizes the ego node’s distinctiveness within the subgraph but also effectively captures the collective characteristics of its surrounding neighbors. 

\subsubsection{Adaptive Feature Scaling Mechanism.} After the above operations, the final features retain richer local and structural information, enhancing the model’s capacity to capture node relationships. However, the reliance on structural information varies across datasets. In some cases, node features alone can achieve results comparable to GNN-based approaches \cite{liu2021non}. Therefore, it is necessary to design a mechanism that adaptively adjusts the importance of different features.

In our framework, after applying differentiated zero-vector padding and the ego-alter representation method, the final representation can be partitioned into four groups based on their original dimensions: \textit{ego-left}, \textit{ego-right}, \textit{pool-left}, and \textit{pool-right}.For Each feature group, we apply a learnable scaling factor, denoted as \(\alpha_1, \alpha_2, \alpha_3, \alpha_4\). These scaling factors are constrained to sum to 1, determined through a softmax operation, and learned with an independent learning rate. In conclusion, The final representation is computed as:

\begin{equation}
h_{\text{final}} = \alpha_1 \cdot h_{\text{ego-left}} \parallel \alpha_2 \cdot h_{\text{ego-right}} \parallel \alpha_3 \cdot h_{\text{pool-left}} \parallel \alpha_4 \cdot h_{\text{pool-right}}
\end{equation}

The aforementioned feature groups exhibit distinct characteristic tendencies: \textit{ego-based features} emphasize the intrinsic characteristics of the \textit{ego node}, while \textit{pool-based features} capture the global structural properties of the subgraph. The distinction between \textit{left} and \textit{right} arises from differences in zero-padding positions. During message propagation, for the \textit{ego node}, \textit{right} acts as the primary channel for receiving information from its neighbors, whereas for \textit{alter nodes}, \textit{left} serves as the main pathway for integrating the ego node’s features. Consequently, these feature groups exhibit different directional tendencies in message propagation.

Finally, the resulting representation is passed through a feedforward neural network (FFNN) to obtain the final predicted label.

\section{Experiment}
\subsection{Setup}
\paragraph{Datasets.} Our experiments are conducted on six widely-used benchmark datasets, including three homophilic graphs: Cora, CiteSeer, and PubMed \cite{kipf2016semi}, as well as three heterophilic graphs: Chameleon, Squirrel \cite{rozemberczki2021multi}, and Film \cite{pei2020geom}. For each dataset, we apply the same data split used in Geo-GCN \cite{pei2020geom}, with 48\%, 32\%, and 20\% of the nodes allocated for training, validation, and testing, respectively\footnote{ Although the paper claims a split ratio of 60\%/20\%/20\%, the actual split shared on GitHub is 48\%/32\%/20\%}. 

\paragraph{Baselines.} Since methods that transform nodes into subgraphs for prediction are primarily explored within the pretraining domain, direct comparisons with these methods may not be entirely fair. To address this, we have developed a baseline approach based on established research. To ensure a fair comparison, we standardize the subgraph construction process and employ a GIN model for learning node representations, but for the final subgraph representation, it is obtained by pooling the features of all nodes, aligning with the most prevalent practices in the literature. We call this approach SubGNN-Base.

On the other hand, to illustrate our model's performance, we also compare our approach with following approachs: 1) Classic GNN models, vanilla GCN \cite{kipf2016semi}, GAT \cite{velivckovic2017graph}, GraphSAGE \cite{hamilton2017inductive} 2) Heterophilic state-of-the-art models, MixHop \cite{abu2019mixhop}, GCNII \cite{chen2020simple}, FSGNN \cite{maurya2022simplifying}, H2GNN \cite{zhu2020beyond}, ACM-GCN \cite{luan2022revisiting}, Gradient Gating \cite{rusch2023gradient}, GloGNN++ \cite{li2022finding}, GPR-GNN \cite{chien2020adaptive}. 3) Multilayer Perceptron (MLP). It's important to note that all of these methods adopt a node-centric perspective.

\paragraph{Implementation Details,} In our experiments, we used cross-entropy loss for classification and Adam as the optimizer, setting a maximum of 150 epochs with early stopping to improve efficiency. We included key hyperparameters such as learning rate, weight decay, dropout ratio, and hidden size as optimization targets. To automate this process, we used Optuna \cite{akiba2019optuna}, which supports intelligent algorithms like Bayesian optimization and genetic algorithms to efficiently explore large hyperparameter spaces. Optuna’s pruning mechanism also accelerates the process by discarding unpromising trials early, enabling faster and more accurate tuning. For each dataset, we performed 150 iterations with Optuna's default TPE (Tree-structured Parzen Estimator) algorithm. Our method was implemented in Python and PyTorch Geometric (PyG) \cite{fey2019fast}, and all experiments ran on an NVIDIA GeForce RTX 4090 GPU with 24GB memory.
\begin{table*}[t]
    \centering
    \caption{Performance comparison of various models on different datasets. The best results are highlighted in bold format, while the second-best results are underlined. N/A indicates that the results are not available in the referenced paper.}
    \begin{adjustbox}{width=\textwidth}
    \begin{tabular}{lccccccc}
        \toprule
        \textbf{Model/Dataset} & \multicolumn{1}{c}{\textbf{Cora}} & \multicolumn{1}{c}{\textbf{CiteSeer}} & \multicolumn{1}{c}{\textbf{PubMed}} & \multicolumn{1}{c}{\textbf{Chameleon}} & \multicolumn{1}{c}{\textbf{Squirrel}} & \multicolumn{1}{c}{\textbf{Film}} \\
        \midrule
        MLP          & 74.75 & 74.02 & 87.16 & 46.21 & 28.77 & 36.53 \\
        GCN          & 87.28 & 76.68 & 87.38 & 59.82 & 36.89 & 30.26 \\
        GAT          & 82.68 & 75.46 & 84.68 & 54.69 & 30.62 & 26.28 \\
        GraphSage    & 86.90 & 76.04 & 88.45 & 58.73 & 41.61 & 34.23 \\
        \midrule
        MixHop       & 87.61 & 76.26 & 85.31 & 60.50 & 60.50 & 32.22 \\
        GCNII        & \underline{88.01} & 77.13 & 90.30 & 62.48 & N/A & N/A \\
        FSGNN        & 87.73 & \underline{77.19} & \underline{89.73} & \underline{78.14} & \underline{73.48} & 35.67 \\
        H2GNN        & 86.92 & 77.07 & 89.40 & 57.11 & 36.42 & 35.86 \\
        ACM-GCN      & 87.91 & \textbf{77.32} & \textbf{90.00} & 54.40 & 54.40 & 36.28 \\
        GRAD. GATING & N/A & N/A & N/A & 71.40 & 64.26 & 37.14 \\
        GloGNN++     & \textbf{88.33} & 77.22 & 89.24 & 71.21 & 57.88 & \textbf{37.70} \\
        GPR-GNN      & 87.95 & 77.13 & 87.54 & 46.58 & 31.61 & 34.63 \\
        \midrule
        SubGNN-Base  & 85.71 & 73.57 & 86.41 & 64.47 & 50.34 & 31.18 \\
        \textbf{SubGND}       & \textbf{88.33} & 77.08 & 89.26 & \textbf{78.68} & \textbf{74.77} & \underline{37.46} \\
        \bottomrule
    \end{tabular}
    \end{adjustbox}
    
    \label{tab:performance_comparison}
\end{table*}
\subsection{Results \& Analysis} 
For all datasets, we report the mean test accuracy across 10 experimental runs. Table \ref{tab:performance_comparison} presents the results for SubGND, alongside other comparative models. The results for MLP, ACM-GCN and GPR-GNN are from \cite{li2022finding}, those for GCN, GAT, GraphSAGE, MixHop are from \cite{zhu2020beyond}, GCNII is from \cite{maurya2022simplifying}, and the results for Gradient Gating, H2GNN, FSGNN, GloGNN++ are from the original papers.

\paragraph{The Performance of SubGNN-Base.} 
As shown in table \ref{tab:performance_comparison}, the results of SubGNN-Base, which constructs the induced graph and directly uses the entire graph representation to predict the corresponding label looks like Interesting. On the one hand, it shows that this method doesn't significantly lag behind most models in homophilic graphs,which is understandable, as in homophilic graphs, most connected nodes tend to form communities that share the same label. Therefore, it is understandable that using the overall representation of the induced subgraph as a substitute for the ego-vertex representation does not perform too poorly.

The results on heterophilic graphs are notable, especially for Chameleon and Squirrel. This method not only outperforms all classic models but also surpasses over half of the GNNs optimized for heterophilic graphs. This suggests that in some datasets, leveraging shared information between a node and its neighbors can be highly effective. Similar findings were reported by \cite{mao2024demystifying}, attributing this success to distinguishable structural patterns. However, like our observations, their study also found that these patterns are not always reliable, as seen in the Film dataset, where performance lags behind many models, with only a slight improvement over GCN and GAT.
\begin{table*}[t]
\centering
\caption{Ablation study on different components in SubGND. The best results are highlighted in bold format}
\renewcommand{\arraystretch}{1.2} % Adjust the value to increase spacing
\begin{adjustbox}{width=0.8\textwidth}
\begin{tabular}{lcccccc}
\toprule
\textbf{Model/Dataset} & \textbf{Cora} & \textbf{Citeseer} & \textbf{Pubmed} & \textbf{Chameleon} & \textbf{Squirrel} & \textbf{film} \\ 
\midrule
SubGND-NoDZP & 83.76 & 74.07 & 87.66 & 77.85 & 72.75 & 29.51 \\
SubGND-NoEAS & 84.77 & 74.37 & 87.91 & 59.80 & 46.94 & 35.98 \\
SubGND-NoScale & 85.70 & 73.77 & 87.84 & 68.42 & 48.57 & 35.70 \\
\textbf{SubGND}   & \textbf{88.33} & \textbf{77.08} & \textbf{89.26} & \textbf{78.68} & \textbf{74.77} & \textbf{37.46} \\
\bottomrule
\end{tabular}
\end{adjustbox}
\label{tab:variant}
\end{table*}

\paragraph{The Performance of SubGND.} As shown in Table 1, our method achieves competitive performance across all benchmarks when compared to the aforementioned models. Specifically, it yields substantial improvements over baseline subgraph-based prediction methods on all datasets, with the largest gain observed on Squirrel, where performance increases by 25 points. 

Compared to node-centric models, SubGND achieves competitive performance across various graph types. On homophilous graphs, it consistently outperforms many models, including all classical methods, and lags behind the state-of-the-art by no more than 0.8 points. On heterophilous graphs, where many models experience significant performance degradation, SubGND achieves either the top or second-best results across all datasets. Notably, This high level of performance is achieved without the need for specialized adaptations to account for heterophily, demonstrating SubGND’s versatility. These results highlight the model's generalizability and the considerable potential it holds for both homophilous and heterophilous graph tasks.

\subsection{Ablation Study.}
In this section, we examine the impact of various components within the proposed framework. Table \ref{tab:variant} presents the experimental results for the proposed model and its variants. Specifically, SubGND-NoDZP removes the differentiated zero-padding method and instead using the original features. SubGND-NoEAS does not utilize the Ego-Alter subgraph representation, relying on the pooled representations of all nodes as the subgraph representation. SubGND-NoScale omits the use of scaling mechanism to adjust the features. Aside from these differences, all other components remain consistent with SubGND.

\paragraph{Results and Analysis.} The results demonstrate that the primary impact of our proposed modules is observed in improving performance on heterophilic graphs, whereas for homophilic graphs, the removal of any single module results in only a modest performance decline, with the most significant difference being approximately 4 points on Cora, and the smallest difference being less than 1 point on PubMed. This observation is intuitive, as nodes in homophilic graphs are highly similar to their neighbors, implying that the difference between using the node’s own features versus using the entire induced subgraph representation is not as pronounced. Consequently, transforming node classification into a subgraph classification task is relatively straightforward for homophilic graphs.

However, in heterophilic graphs, the presence of label conflicts may lead to performance degradation due to the indiscriminate incorporation of neighboring node information, which can obscure the node’s intrinsic features. This necessitates a more nuanced approach that distinguishes between the ego-vertex and alter-vertices. In Chameleon and Squirrel datasets, for instance, the removal of either the Ego-Alter subgraph representation mechanism or the use of a scaling mechanism leads to a substantial drop in performance, with reductions of up to 28 percentage points. Both of these methods are specifically designed to mitigate the issues that arise from label conflicts. Furthermore, for the Film dataset, the differentiated zero-padding strategy proves to be a crucial module. We hypothesize that this is due to the dataset's higher dependency on the node's intrinsic features, and the primary goal of the strategy is to reduce the confusion between the node's own information and that of other nodes.

% \begin{table}[t]
% \centering
% \caption{ The learned scaling factors (\(\alpha\)-values) under optimal conditions across different datasets}
% \renewcommand{\arraystretch}{1.2} % Adjust the value to increase spacing
% \begin{adjustbox}{width=0.8\columnwidth}
% \begin{tabular}{|l|cccccc|}
% \hline
% \textbf{$\alpha$-values/Dataset} & \textbf{Cora} & \textbf{Citeseer} & \textbf{Pubmed} & \textbf{Chameleon} & \textbf{Squirrel} & \textbf{Film} \\ 
% \hline
% \textbf{\(\alpha_1\) (\(h_{\text{ego-left}}\))} & 0.261  & 0.302 & 0.178 & 0.387 & 0.244 & 0.177 \\
% \textbf{\(\alpha_2\) (\(h_{\text{ego-right}}\))} & 0.263 & 0.310 & 0.333 & 0.376 & 0.484 & 0.440  \\
% \textbf{\(\alpha_3\) (\(h_{\text{pool-left}}\))} & 0.239 & 0.294 & 0.341 & 0.125 & 0.119 & 0.197 \\
% \textbf{\(\alpha_4\) (\(h_{\text{pool-right}}\))} & 0.237 & 0.094 & 0.148 & 0.112 & 0.154 & 0.186 \\
% \hline
% \end{tabular}t 
% \end{adjustbox}
% \label{tab:alpha_values}
% \end{table}

\begin{table}[t]
\centering
\caption{The learned scaling factors under optimal conditions across different datasets}
\renewcommand{\arraystretch}{1.3} 
\begin{tabular}{ccccccc}
\toprule
$\alpha$-values/Dataset & Cora & Citeseer & Pubmed & Chameleon & Squirrel & Film \\
\hline
\textbf{\(\alpha_1\) (\(h_{\text{ego-left}}\))} & 0.261  & 0.302 & 0.178 & 0.387 & 0.244 & 0.177 \\
\textbf{\(\alpha_2\) (\(h_{\text{ego-right}}\))} & 0.263 & 0.310 & 0.333 & 0.376 & 0.484 & 0.440  \\
\textbf{\(\alpha_3\) (\(h_{\text{pool-left}}\))} & 0.239 & 0.294 & 0.341 & 0.125 & 0.119 & 0.197 \\
\textbf{\(\alpha_4\) (\(h_{\text{pool-right}}\))} & 0.237 & 0.094 & 0.148 & 0.112 & 0.154 & 0.186 \\
\bottomrule
\end{tabular}
\label{tab:alpha_values}
\end{table}
\paragraph{Feature Importance Analysis.} To evaluate the relative importance of different feature groups within our model, we examined the learned scaling factors (\(\alpha\)-values) under optimal conditions. Higher \(\alpha\)-values indicate greater significance, whereas lower \(\alpha\)-values signify lesser importance. 

\begin{figure*}[t]
\centering
\includegraphics[width=0.9\textwidth]{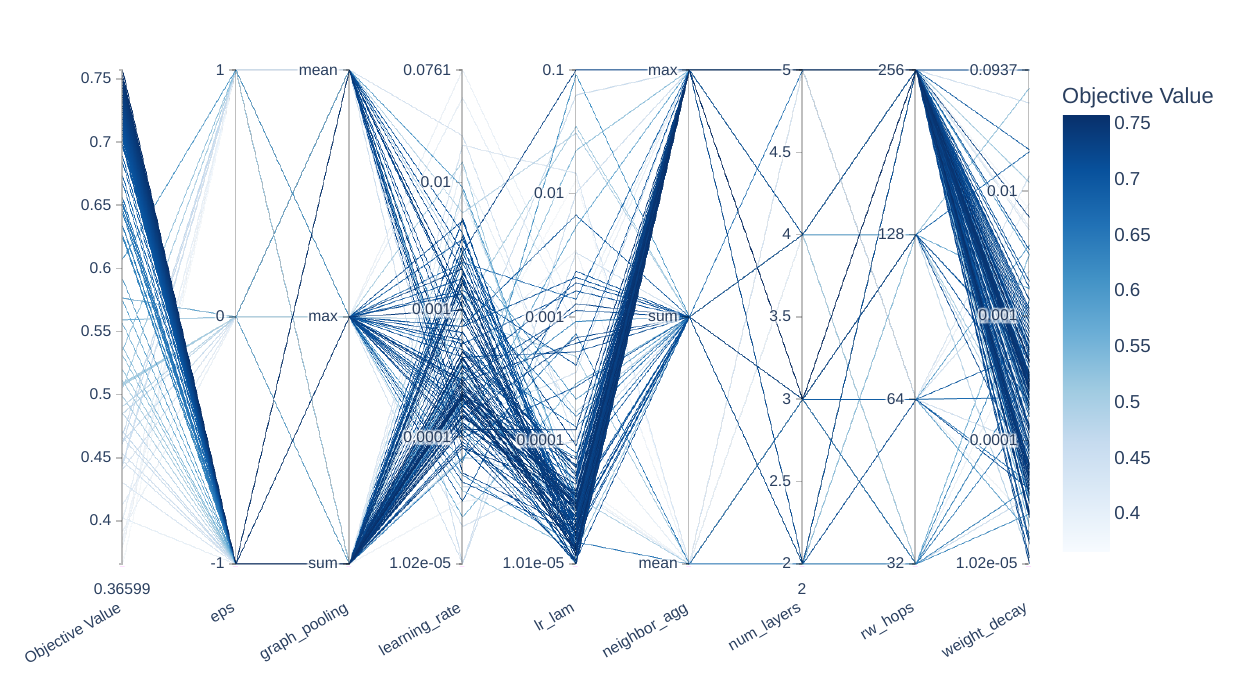} 
\caption{Parallel Coordinate Plot for Hyperparameter Search in Squirrel Dataset}
\label{fig3}
\end{figure*}

As illustrated in Table \ref{tab:alpha_values}, feature preferences vary even among homophilic graphs. For example, in the Cora dataset, ego-vertex features and pooled alter-vertex features are almost equally important. Conversely, Citeseer prioritizes ego-vertex features, whereas PubMed favors the interaction between the latter part of the ego-vertex features and the initial part of the pooled alter-vertex features, highlighting the model's emphasis on feature interactions. In heterophilic graphs, the model places greater importance on ego-vertex features, suggesting that incorporating features from closely related nodes might introduce noise. These findings imply that using a fully induced graph representation to distinguish between ego-vertices may be suboptimal.
\paragraph{Parameter Analysis.} To further evaluate the effectiveness and adaptability of our framework across different datasets, we utilize parallel coordinate plots to visualize hyperparameter variations during the search process. As an illustrative example, we present the results from the \textbf{Squirrel} dataset (see Figure \ref{fig3}).
  
From the plot, we observe that \(\text{eps} = -1\) is particularly significant for this dataset (this also holds for Chameleon). This parameter regulates the balance between self-features and neighbor features during message passing, where \(\text{eps} = -1\) indicates that neighbor distributions play a more discriminative role in node classification (see Equation \ref{eq2}, where \(-1\) excludes self-features and \(+1\) amplifies their influence). Additionally, larger \(\text{rw\_hops}\) values generally lead to improved performance, whereas hyperparameters such as \(\text{num\_layers}\) and \(\text{learning rate}\) exhibit minimal impact. 

It is important to note that these observations are specific to the Squirrel dataset, and different datasets demonstrate distinct behaviors. For instance, in most homophilic graphs, the \(\text{eps}\) parameter has little influence, while in Pubmed, increasing \(\text{rw\_hops}\) does not consistently enhance performance — a moderate value (128) is optimal. These findings reinforce that optimal hyperparameter configurations vary significantly across datasets, highlighting the necessity of adaptive mechanisms in model design to ensure robustness and generalization.

\paragraph{Efficiency and Scalability.} The proposed method introduces additional computational overhead during preprocessing, primarily due to induced subgraph generation. However, this is a one-time cost incurred only at the initial training stage, with its computational complexity dependent on the sampling strategy. In terms of scalability, global message-passing GNNs struggle with large-scale datasets due to high computational and memory demands, whereas the proposed framework maintains comparable complexity to existing subgraph sampling methods while achieving superior performance.

\section{Conclusion} This work proposes SubGND, a novel framework that reformulates node classification as a subgraph classification task. A key challenge in this transformation is \textit{label conflict}, where nodes with identical neighboring subgraphs require different labels due to type differences, leading to confounding information that degrades model performance. To address this, we introduce a Differentiated Zero-Padding strategy and an Ego-Alter subgraph representation to resolve label conflicts and enhance expressiveness. Additionally, an Adaptive Feature Scaling Mechanism dynamically adjusts feature contributions based on dataset-specific dependencies. Experimental results show that SubGND performs on par with global message-passing GNNs in homophilic graphs while significantly outperforming existing methods in heterophilic settings, demonstrating its effectiveness.

\begin{credits}
% \subsubsection{\ackname} A bold run-in heading in small font size at the end of the paper is
% used for general acknowledgments, for example: This study was funded

\subsubsection{\discintname}
% It is now necessary to declare any competing interests or to specifically
% state that the authors have no competing interests. Please place the
% statement with a bold run-in heading in small font size beneath the
% (optional) acknowledgments,
% for example: 
The authors have no competing interests to declare that are
relevant to the content of this article. 
% Or: Author A has received research
% grants from Company W. Author B has received a speaker honorarium from
% Company X and owns stock in Company Y. Author C is a member of committee Z.
\end{credits}

\bibliographystyle{splncs04}
% \bibliography{ECML_PKDD_2025_Author_Kit/ref}
% \bibliography{ref}
%% Note that this preceding line implies that you store your BibTeX references in a file called 'mybibliography.bib'. If you instead store your references in a file with a different name, for instance 'references.bib', the preceding line should read '\bibliography{references}'. Whatever you do, DO NOT put the file name extension .bib inside the \bibliography command; this will trip up LaTeX compilers. 
%
% If you do not want to use BibTeX, you can also type up the bibliography exactly as you see fit, using the following structure:
% % Note that this number 8 reserves an amount of space (equal to the natural width of the given number) for the label of your references; if you have more than 9 references, you will want to change this number to 18. If you have more than 19 references, this number is best changed to 88. If you have more than 99 references, I salute you.




% LNCS Homepage, \url{http://www.springer.com/lncs}, last accessed 2023/10/25
% \end{thebibliography}
\end{document}